\def\year{2022}\relax
\documentclass[letterpaper]{article} 
\usepackage{aaai22}  
\usepackage{times}  
\usepackage{helvet}  
\usepackage{courier}  
\usepackage[hyphens]{url}  
\usepackage{graphicx} 
 \usepackage{color}
\usepackage{natbib}  
\usepackage{caption} 
\DeclareCaptionStyle{ruled}{labelfont=normalfont,labelsep=colon,strut=off} 
\usepackage{algorithm}
\usepackage{algorithmic}
\usepackage{booktabs}

\newtheorem{definition}{Definition}

\usepackage{amsmath, amssymb}
\usepackage{bm}
\usepackage{subcaption}
\usepackage{multirow}
\usepackage{newfloat}
\usepackage{listings}
\floatstyle{ruled}
\newfloat{listing}{tb}{lst}{}
\floatname{listing}{Listing}
\usepackage{cancel} 
\newcommand{\mbf}[1]{\mathbf{#1}}

\newcommand{\wemb}[1]{$\overrightarrow{\textit{{#1}}}$}

\title{Word Embeddings via Causal Inference: Gender Bias Reducing and Semantic Information Preserving}
\author{
    Lei Ding\textsuperscript{\rm 1}, Dengdeng Yu\textsuperscript{\rm 3}, Jinhan Xie\textsuperscript{\rm 1},Wenxing Guo\textsuperscript{\rm 1}, Shenggang Hu\textsuperscript{\rm 2}, Meichen Liu\textsuperscript{\rm 1}, \\ Linglong Kong\textsuperscript{\rm 1}*, Hongsheng Dai\textsuperscript{\rm 2}, Yanchun Bao\textsuperscript{\rm 2}, Bei Jiang\textsuperscript{\rm 1}\\
    
}
\affiliations{
    \textsuperscript{\rm 1}Department of Mathematical and Statistical Sciences, University of Alberta\\

    \{lding1,jinhan3,wenxing2,meichen,lkong,bei1\}@ualberta.ca\\
    \textsuperscript{\rm 2}Department of Mathematical Sciences,
    University of Essex\\
    \{sh19509, hdaia, ybaoa\}@essex.ac.uk \\
    \textsuperscript{\rm 3}
    Department of Mathematics,
    University of Texas at Arlington\\
    \{dengdeng.yu\}@uta.edu\\

}

\usepackage{bibentry}

\begin{document}

\maketitle

\begin{abstract}

With widening deployments of natural language processing (NLP) in daily life, inherited social biases from NLP models have become more severe and problematic. Previous studies have shown that word embeddings trained on human-generated corpora have strong gender biases that can produce discriminative results in downstream tasks. 
Previous debiasing methods focus mainly on modeling bias and only implicitly consider semantic information while completely overlooking the complex underlying causal structure among bias and semantic components. To address these issues, we propose a novel methodology that leverages a causal inference framework to effectively remove gender bias.  The proposed method allows us to construct and analyze the complex causal mechanisms facilitating gender information flow while retaining oracle semantic information within word embeddings. Our comprehensive experiments show that the proposed method achieves state-of-the-art results in gender-debiasing tasks. In addition, our methods yield better performance in word similarity evaluation and various extrinsic downstream NLP tasks.

\end{abstract}

\setcounter{section}{1}
\section{Introduction}

Word embeddings are dense vector representations of words trained from human-generated corpora \citep{mikolov2013efficient,pennington2014glove}. Word embeddings have become an essential part of natural language processing (NLP). However, it has been shown that stereotypical bias can be passed from human-generated corpora to word embeddings \citep{weatscience,rnd,zhao2019gender}. 

With wide applications of NLP systems to real life, biased word embeddings have the potential to aggravate and possibly cause serious social problems. For example, translating `He is a nurse' to Hungarian and back to English results in `She is a nurse' \citep{douglas2017ai}. In word analogy tasks appears in \citet{genderpca}, wherein \wemb{she} is closer to \wemb{nurse} than \wemb{he} is to \wemb{doctor}. \citet{zhao2018learning} shows that biased embeddings can lead to gender-biased identification outcomes in co-reference resolution systems.

Current studies on word embedding bias reductions can be divided into two camps: word vector learning methods \citep{zhao2018learning} and post-processing algorithms \citep{genderpca,kaneko2019gender}. Word vector learning methods are time-consuming and suffer from the high computational cost required to train word embeddings from scratch. 
To overcome these limitations, post-processing algorithms have emerged as popular alternatives. \citet{yang2020causal}, for example, proposes a simple and efficient algorithm that projects embeddings into a space that is orthogonal to gender-specific words such as \emph{mother} and \emph{father} and is successful in reducing gender bias. 

However, the critical issue of using gender-specific word vectors remains: 
information on gender and semantics entangled within these words. For example, the gendered word pair \emph{bride} and  \emph{bridegroom} exhibits gender information as well as semantic information pertaining to weddings.
Therefore, eliminating gender information through pairs of gendered words such as 
\emph{policeman} and \emph{policewoman} or \emph{wizard} and \emph{witch},
also eliminates intrinsic semantic information: this is clearly not ideal. 

As a solution, we propose utilizing the differences between vectors corresponding to paired gender-specific words to better eliminate gender bias while retaining important semantic information. 
These differences are between embedded vectors for male- and female-gendered words, such as \wemb{father}$-$\wemb{mother} or \wemb{bridegroom}$-$\wemb{bride}.
As a motivating example\footnote{Please refer to the appendix for detail explanation}, Table \ref{tab:subsec:exp1} demonstrates that this simple change from gender-specific word vectors to the differences between word-pair vectors indeed retains more semantic information than the state-of-the-art post-processing framework \citep{yang2020causal}.


\begin{table}[t]
\fontsize{8.5}{10}\selectfont
\centering
\begin{tabular}{@{}rcccc@{}}
\toprule
 &
  \begin{tabular}[c]{@{}c@{}}Task 1\\ \emph{Wedding}\end{tabular} &
  \begin{tabular}[c]{@{}c@{}}Task 2\\ \emph{Service}\end{tabular} &
  \begin{tabular}[c]{@{}c@{}}Task 3\\ \emph{Family} \end{tabular} &
  \begin{tabular}[c]{@{}c@{}}Task 4\\ \emph{Religion} \end{tabular} \\ \midrule
Oracle              & 11.22 (0.20) & 9.96 (0.11) & 13.51 (0.30)& \
20.27 (0.30) \\
DeSIP   & 7.01 (0.15)   & 6.67 (0.10) & 10.69 (0.25) & 13.59 (0.25)   \\
HSR                    & 4.34 (0.14)  & 5.61 (0.10) & 8.90 (0.22)  & 9.85 (0.20)  \\ \midrule
Win-loss & 100.00\%         & 99.00\%        & 100.00    \%     & 100.00\%         \\ \bottomrule
\end{tabular}
\caption{Semantic information preservation experiment. 
}
\label{tab:subsec:exp1}
\end{table}

In this paper, we propose novel causal frameworks for reducing bias in word embeddings while maximally preserving semantic and lexical information.  
Our contributions are summarized as follows.
\begin{itemize}
    \item We develop two causal inference frameworks for reducing biases in word embeddings that improve upon existing state-of-the-art methods. 
    \item We find an intuitive and effective way to better represent gender-related information that needs to be removed and use this approach to achieve oracle-like semantic and lexical information retention.
    \item We show that our methods outperform other \emph{state-of-the-art} debiasing methods in various downstream NLP tasks.
\end{itemize}

The rest of this paper is organized as follows. We first present a thorough review of current studies on word embedding bias evaluation and debiasing algorithms.  We then define two types of bias and propose frameworks for dealing with each. The comprehensive experimental results on a series of gender bias evaluation and semantic evaluation tasks demonstrate the effectiveness of our proposed methods.

\section{Related Works}

\subsection{Quantifying Gender Bias}
Numerous studies have demonstrated that word embeddings trained by human-generated corpora exhibit human stereotype bias. \citet{weatscience} develops the Word Embedding Association Test (WEAT) as an analogue to the Implicit Association Test used in psychology \citep{greenwald1998measuring} to detect implicit stereotypes.
WEAT measures the association between a word and an attribute using cosine similarity; the test compares two sets of target words against a pair of attribute sets. 

\citet{genderpca} applies word analogy tests as a way to demonstrate bias. The task uses a word embedding to find an output to pair with a given input word, say, \emph{doctor}, such that the (target, output) pair is in analogy to the gender pair (he, she).
The word embedding passes the test if the output is stereotype-free, say, \emph{physician} instead \emph{nurse} for the input \emph{doctor}.
However, this task requires crowd-sourcing to set the benchmark and has been replaced by other evaluation methods in more recent works.

Another approach from \citet{genderpca} for evaluating gender bias involves computing projections onto a gender direction, the difference between vector embeddings of a pair of gender-specific words (e.g., he and she, as the most widely accepted definition). This debiasing metric is used in many other studies \citep{multiclass}.
Such a method has failed to become the gold standard because a ``true'' gender direction if it exists, is used in the evaluation.

\citet{gonen} later points out that direct projection does not eliminate gender bias from the geometry of the embedding and that biased words tend to cluster together even after debiasing.
To account for this, the neighborhood bias metric was introduced to measure the bias of a word by counting the difference in the number of (socially) male- and female-biased neighbors among the word's $K$-nearest neighbors.


\subsection{Prior Debiasing Methods}
Current studies on word embedding bias reductions can be divided into two camps: word vector learning methods \citep{zhao2018learning} and post-processing algorithms  for instance \cite{genderpca} and \cite{kaneko2019gender} and many more. 
Word vector learning methods require retraining of the word embedding and can be time-consuming due to the retraining of the word embedding.
Therefore, most of the works on debiasing word embeddings choose to remove the bias through post-processing, including algorithms like \citep{genderpca, kaneko2019gender, dev2019attenuating, wang2020double, shin2020neutralizing,yang2020causal}.

From a technical perspective, we see that \citet{genderpca} formulates the core idea of detecting the subspace that contains the most information related to gender
Based on the idea of removing gender subspace, other works have incorporated different strategies, e.g., maximizing the distance between masculine and feminine words \cite{zhao2018learning}, detecting gender direction using partial projection \cite{dev2019attenuating}, or detecting and mitigating distortion in gender direction due to word frequency \cite{wang2020double}.
Various extensions of \citep{genderpca} are also developed, for instance removing bias with respect to multiclass attributes (like ethnic) \citep{multiclass} or debiasing multilingual word embeddings \cite{bansal2021debiasing}.

More recent works \citep{yang2020causal, shin2020neutralizing} have considered the problem beyond just detecting and removing gender direction from gender-neutral word vectors.
\citet{shin2020neutralizing} models a word vector as a sum of two components, each containing latent gender information and semantic information respectively. 
An autoencoder is trained to disentangle these two components and gender-neutral words are debiased using a counterfactual copy of itself, i.e. a synthesized word vector with the same semantic component but biased in the other gender direction.

Similarly, \citet{yang2020causal} approaches the problem using a causal framework in which it is assumed that latent gender information affects both gendered and gender-biased words. The model aims to recover gender-specific information in gender-biased words from the gendered words through a linear ridge regression.
In comparison, the causal framework used in our approach not only distinguishes gender information from semantic information but also takes into account the potential effect of the former on the latter through causal inference.
This causal path from gender information to semantic information is overlooked by the causal model used in \citep{yang2020causal}.

\section{Methodology}

\subsection{Preliminary Definitions}
We characterize two types of gender bias in the causal framework and propose algorithms 
for removing each type. Specifically, we use model intervention techniques to determine causal effects in a causal model. It is more manageable to apply the model intervention to proxy variables of the gender bias rather than the gender bias variables themselves (represented by the differences between gender-specific word pair vectors, such as \wemb{he}$-$\wemb{she} or \wemb{male}$-$\wemb{female}), since the latter are generally regarded as inherited attributes for which interventions are often impossible in practice.

We consider five types of variables corresponding to five word-related matrices: an $s_1$-dimensional pure gender bias variable $D$ with a corresponding matrix $\mbf{D} \in \mathcal{R}^{N \times s_1}$ composed of pure gender bias vectors such as \wemb{he}$-$\wemb{she} and \wemb{male}$-$\wemb{female}; 
an $s_2$-dimensional gender bias variable proxy $P$ with a corresponding matrix $\mbf{P} \in \mathcal{R}^{N \times s_2}$
composed of vectors that are directly influenced by $D$ that should not affect the final prediction; an $m$-dimensional resolving, non-gender-specific word variable $Z$ with a corresponding matrix $\mbf{Z}\in\mathcal{R}^{N \times m}$ composed of vectors that are influenced by $D$ in a manner that we accept as non-discriminatory;
a $d$-dimensional, non-gender-specific word variable $Y$ with a corresponding matrix $\mbf{Y}\in \mathcal{R}^{N \times d}$
composed of word vectors potentially containing gender bias that needs to be removed, such as
\wemb{nurse} and \wemb{engineer}; and another $p$-dimensional, non-gender-specific word variable $X$ with a corresponding matrix $\mbf{X} \in \mathcal{R}^{N \times p}$ that may retain  semantic information. Here $N$ is the dimension of the word embedding vector, and $s_1$, $s_2$, $m$, $d$, and $p$ are the sizes of the variables $D$, $P$, $Z$, $Y$ and $X$, respectively.

It is clear that using the vectors in $\mbf{D}$ can eliminate pure gender bias information contained in word embeddings. In this way, semantic information can be preserved. As shown in Figures \ref{fig:proxy} and \ref{fig:unresolved}, we generally allow influence along the pathway $D\rightarrow X\rightarrow Y$ in our framework. 
Motivated by \citet{kilbertus2017avoiding} and these conventions, we introduce the following definitions.

\begin{definition} (Potential proxy bias.) A variable $Y$ in a causal graph exhibits potential proxy bias if there exists a directed path from $D$ to $Y$ that is blocked by a proxy variable $P$ and if \  $ Y$ itself is not a proxy. 
\end{definition}

This definition indicates that potential proxy bias from $P$ articulates a causal criterion that is in a sense dual to unresolved bias from $Z$.

\begin{definition} (Unresolved bias.) A variable $Y$ in a causal graph exhibits unresolved bias if there exists a directed path from $D$ to $Y$ that is not blocked by a resolving variable $Z$ and $Y$ itself is non-resolving.
\end{definition}

This definition implies that all paths from a gender-bias variable $D$ are problematic unless they are justified by a resolving variable $Z$.

    \subsection{Removing Potential Proxy Bias}

 We now develop a practical procedure for removing proxy bias in a linear structural equation model. 
For each $\mbf{y}^{\rm{}} \in \mathcal{R}^{N}$, the column vector of $\mbf{Y}$, it can be decomposed into two parts as $\mbf{y}^{\rm{}}  = \mbf{y}_{\Delta} + \mbf{y}_{\Delta^\perp}$, where $\mbf{y}_{\Delta}$ and $\mbf{y}_{\Delta^\perp}$ are the projections of $ \mbf{y}^{\rm{}}$ onto the mutually orthogonal spaces $\Delta$ and $\Delta^\perp$. In particular, let $\bm\phi_j \in \mathcal{R}^N$ denote the basis vectors for $\Delta $ and $\bm\psi_{j^\prime} \in \mathcal{R}^N$ denote the basis vectors for $\Delta^\perp $. The whole space $\Omega = \Delta \cup \Delta^\perp$. We can write $ \mbf{y}^{\rm{}}  = \sum_{j: \bm\phi_j \in \Delta} \xi_j \bm\phi_j + \sum_{j^\prime: \bm\psi_{j^\prime} \in \Delta^\perp} \kappa_{j^\prime} \bm\psi_{j^\prime}$, where $\xi_j, \kappa_{j^\prime} \in \mathcal{R}$. In this paper, we take $\Delta =\mathrm{Span}(\mbf{D})$, namely, the linear space spanned by the column vectors of $\mbf{D}$. Consequently, $\Delta^\perp$ contains the semantic information not described by $\mbf{D}$. 
As bias reduction is primarily concerned with reducing bias along paths starting from $D$, we do not remove information from $\mbf{y}_{\Delta^\perp}$. 
\begin{figure}[htbp]
\captionsetup[subfigure]{justification=centering}
\centering
 \subcaptionbox{Proxy bias}[0.45\linewidth]
 {\includegraphics[scale = 0.14]{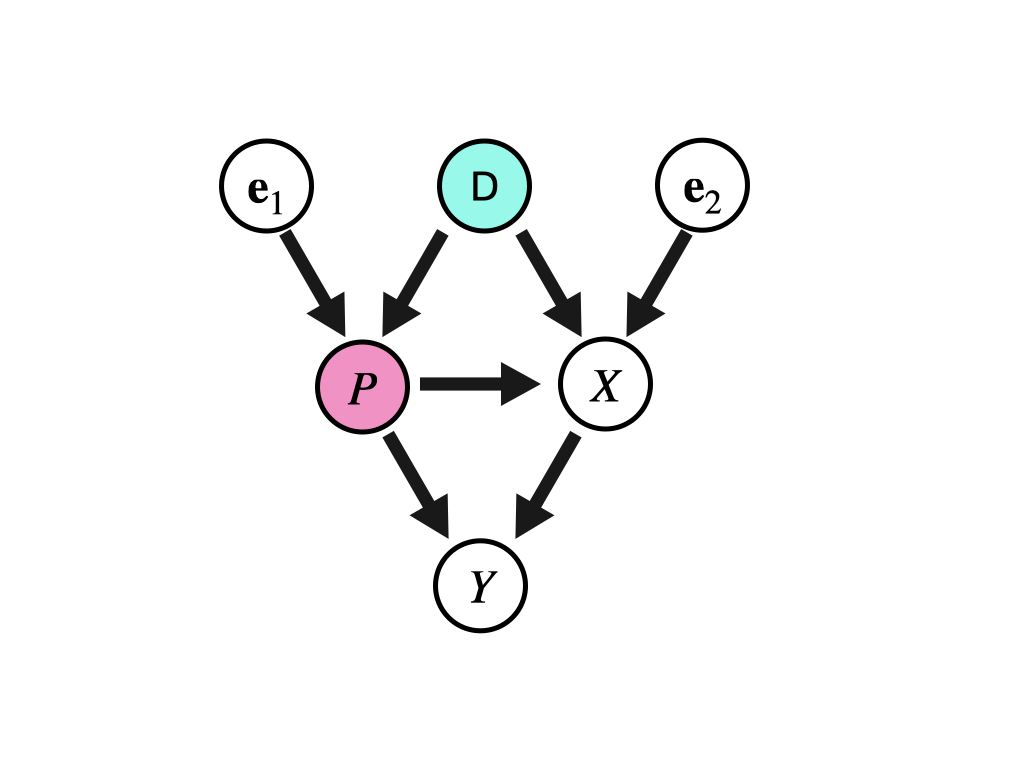}}
 \subcaptionbox{Intervention on proxy bias}[0.45\linewidth]
 {\includegraphics[scale = 0.14]{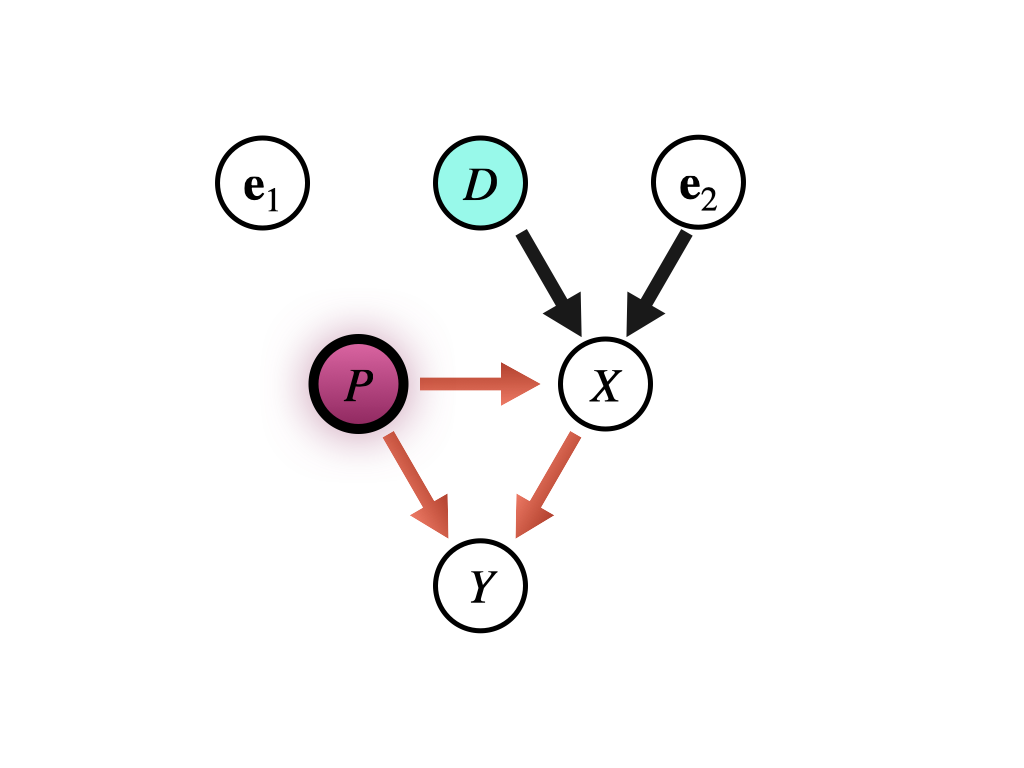}}
\caption{A causal graph for proxy bias removal.}
\label{fig:proxy}
\end{figure}

We next propose an algorithm for debiasing non-gender-specific word vectors $\mbf{y}$. As illustrated in Figure \ref{fig:proxy}, the corresponding linear structural equations are
\begin{align}
\mbf{P} &= \mbf{D} \bm\alpha_0 + \bm{e}_1\nonumber\\ 
\mbf{X} &= \mbf{D} \bm\alpha_1 + \mbf{P} \bm\alpha_2  + \bm{e}_2\\
\mbf{Y} &= \mbf{P}\bm\beta_1 + \mbf{X}\bm\beta_2,\nonumber    
\end{align}
where $\bm{e}_1$ and $\bm{e}_2$ are unobserved errors and $\bm\alpha_0\in\mathcal{R}^{s_1\times s_2}$, $\bm\alpha_1\in\mathcal{R}^{s_1\times p}$, $\bm\alpha_2\in\mathcal{R}^{s_2\times p}$, $\bm\beta_1\in\mathcal{R}^{s_2 \times d}$ and $\bm\beta_2\in\mathcal{R}^{p\times d}$ are parameters. Here, we note that the proxy matrix $\mbf{P}$ contains vectors of words that are direct descendants of $\mbf{D}$ and should not affect the prediction of $\mbf{Y}$. In this paper, we pre-specify $\mbf{P}$ using the gendered-word pairs listed in \citet{zhao2018learning}. We build predictors that remove proxy bias by intervening on $P$, that is, by setting $P=p^\prime$, where $p^\prime$ is a random variable: this is similar to the approach in \citet{kilbertus2017avoiding}. In particular, we want to guarantee that $P$ has no overall influence on the prediction of the non-gender-specific variable $Y$ by adjusting the $P\rightarrow Y$ pathway to cancel the influence along $P\rightarrow X\rightarrow Y$. We do not generally prohibit the potential for the gender bias variable $D$ to influence the non-gender-specific variable $Y$ in this case: see Figure \ref{fig:proxy}.
The non-gender-specific word matrix $\widehat{\mbf{Y}}$ with potential proxy bias removed is\footnote{Please refer to appendix for detail derivation}
\begin{equation}
\widehat{\mbf{Y}} = (\mbf{X} - \mbf{P} \widehat{\bm\alpha}_2 ) \widehat{\bm\beta}_2,  
\end{equation}
where the parameters $\widehat{\bm\alpha}_2$ and $\widehat{\bm\beta}_2$ are estimated by partial least squares (PLS), a supervised dimension reduction method that works particularly well when variable dimensionality is very large \citep{vinzi2012handbook} and becomes a popular tool in various scientific areas in recent years \citep{yu2016partial}.
However, since the debiasing procedure above does not retain any information of $\mbf{Y}_{\Delta^\perp}$ since $\widehat{\mbf{Y}}$ is a descendant of $\mbf{D}$, we must find a way to restore the information of $\mbf{Y}_{\Delta^\perp}$.

In particular, 
we propose obtaining a least-squares estimate $\widehat{\mbf{Y}}_{\Delta}$ of $\mbf{Y}_{\Delta}$ through multivariate linear regression of $\mbf{Y}^{\rm{}}$ on $\mbf{D}$. We then use the residual $\widehat{\mbf{Y}}_{\Delta^\perp} $ as an estimate of $\mbf{Y}_{\Delta^\perp}$. 
Finally, we compute $\widehat{\mbf{Y}}_{\text{P-DeSIP}} = \widehat{\mbf{Y}} + \widehat{\mbf{Y}}_{\Delta^\perp}$ as the bias-reduced version of $\mbf{Y}^{\rm{}}$.
This post-processing algorithm is formally presented in Algorithm 1.


\setcounter{algorithm}{0}
\begin{algorithm}[htb]
\caption{(P-DeSIP) Removing potential proxy bias.}
\label{alg:FramworkP}
\begin{algorithmic}[1] 
\renewcommand{\algorithmicrequire}{{{\bf Input}:}}
\REQUIRE 
$\mbf{D}$, 
$\mbf{P}$, 
$\mbf{X}$, 
and $\mbf{Y}$.
\STATE Solve $\mbf{X} = \mbf{D} \bm\alpha_1 + \mbf{P}\bm\alpha_2  + \bm{e}_2$ by PLS to get ($\widehat{\bm\alpha}_1$, $\widehat{\bm\alpha}_2$)
\STATE Solve $\mbf{Y} =  \mbf{P}\bm\beta_1 + \mbf{X}\bm\beta_2 $ by PLS to get ($\widehat{\bm\beta}_1$, $\widehat{\bm\beta}_2$)
\STATE Compute $\widehat{\mbf{Y}} = (\mbf{X} - \mbf{P} \widehat{\bm\alpha}_2 ) \widehat{\bm\beta}_2$
\STATE Compute $\widehat{\mbf{Y}}_{\Delta^\perp} = \mbf{Y} - \mbf{D} (\mbf{D}^T \mbf{D})^{-1} \mbf{D}^T \mbf{Y}$ 
\STATE Compute $\widehat{\mbf{Y}}_{\text{P-DeSIP}} = \widehat{\mbf{Y}} + \widehat{\mbf{Y}}_{\Delta^\perp}$
\renewcommand{\algorithmicrequire}{{{\bf Output}:}}
\REQUIRE
$\widehat{\mbf{Y}}_{\text{P-DeSIP}}$ as debiased word matrix.
\end{algorithmic}
\end{algorithm}
In practice, when the dimensionality of $\mbf{X}$ is extremely high, the computational cost of this algorithm becomes a concern. With this in mind, we introduce a preliminary screening step to reduce ultrahigh dimensionality to a moderate level before conducting a refined analysis. Before conducting a simple screening procedure using correlation learning, each column of $\mbf{X}$ and $\mbf{Y}$ are standardized to a mean of zero and a standard deviation of one. Inspired by \citet{fan2008sure} and \citet{xie2020category}, we propose the following marginal screening utility to measure the dependence between $\mbf{Y}$ and the columns $\mbf{x}_k$ ($k=1,\dots, p$) of $\mbf{X}$:
$\tau_k = \max\limits_{j=1,\ldots,d}|\mbf{x}_k^{\top}\mbf{y}_j|/N,$
where $\mbf{y}_j$ ($j=1,\ldots,d$) denotes the $j$-th column of $\mbf{Y}$. We propose ranking  $\mbf{x}_k$ by sorting $\tau_k$ from largest to smallest. We denote the reduced non-gender-specific word matrix by $\mbf{X}_{\widehat{\mathcal{M}}},$ where $\widehat{\mathcal{M}} = \{k : \tau_k\geq \gamma_n\}$ and $\gamma_n$ is a pre-specified threshold value.

\subsection{Removing Unresolved Bias}
\begin{figure}[htbp]
\captionsetup[subfigure]{justification=centering}
\centering
 \subcaptionbox{Unresolved bias}[0.45\linewidth]
 {\includegraphics[scale = 0.14]{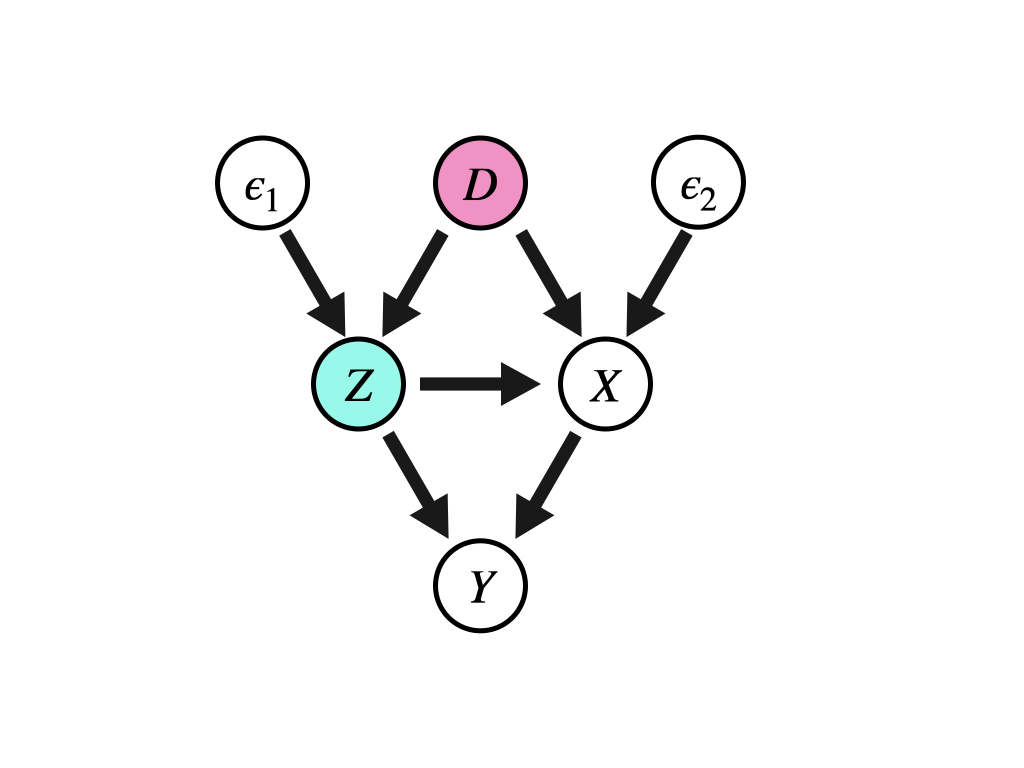}}
 \subcaptionbox{Intervention on unresolved bias}[0.45\linewidth]
 {\includegraphics[scale = 0.14]{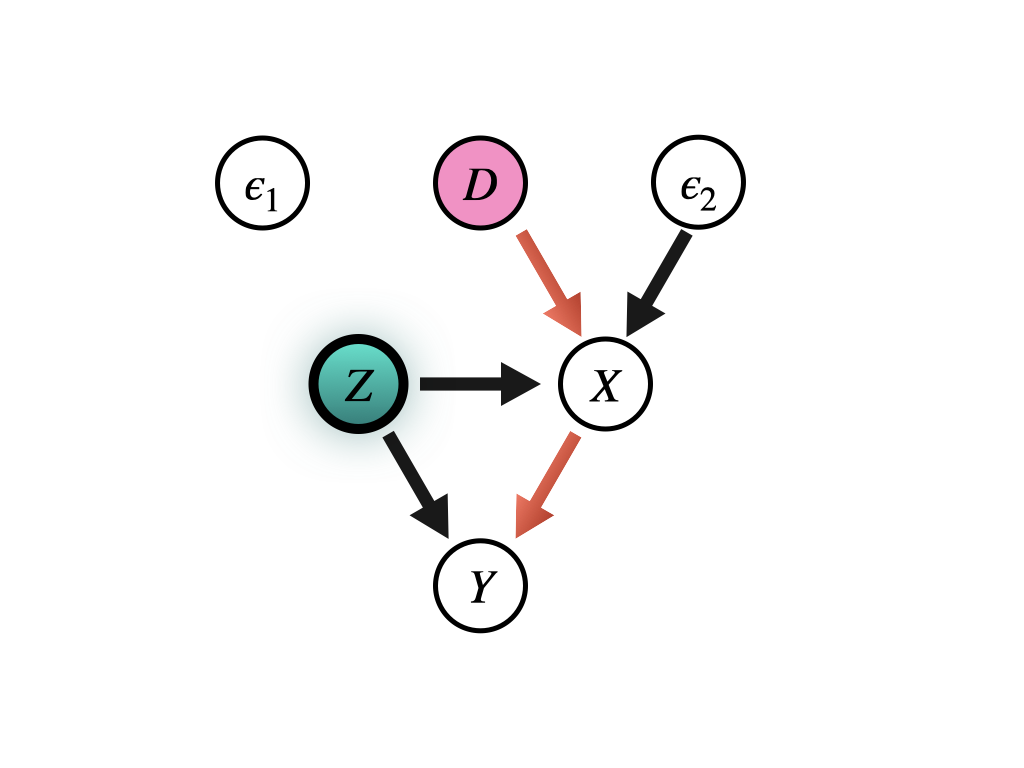}}
\caption{A causal graph for unresolved bias removal. }
\label{fig:unresolved}
\end{figure}

We take a similar approach to remove unresolved bias when a proxy gender bias matrix $\mbf{P}$ is not attainable.
We consider the resolving non-gender-specific word matrix $\mbf{Z}\in\mathcal{R}^{N \times m}$ that directly affects $\mbf{X}$  instead of the proxy bias matrix $\mbf{P}$: this is illustrated in Figure \ref{fig:unresolved}. 

Resolving variables are influenced by $\mbf{D}$ in a manner that we accept as non-discriminatory: therefore, ${\mbf Z}$ is chosen to directly affect $\mbf{X}$ and have some correlation with $\mbf{D}$. In particular, we choose $\mbf{Z}$  containing the adjectives and nouns correlated to $\mbf{D}$ based on mean cosine similarity, while $\mbf{X}$ includes the words that are otherwise contained by $\mbf{Y}$, $\mbf{Z}$, and $\mbf{D}$. Since all adjectives in English have an adverb form, this ensures that the path from $\mbf{Z}$ to $\mbf{X}$ exists.


The causal dependencies in the corresponding linear structural equation model are equivalent to those in Figure \ref{fig:proxy} for potential proxy bias:
\begin{align}
{\mbf Z} &= {\mbf D}{\bm \gamma}_0  + \bm{\epsilon}_1\nonumber\\
{\mbf X} &=  {\mbf D}{\bm \gamma}_1 +  {\mbf Z}{\bm\gamma}_2 + \bm{\epsilon}_2\\ {\mbf Y} &=  {\mbf Z}{\bm \theta}_1 +  {\mbf X}{\bm \theta}_2,\nonumber    
\end{align}
where $\bm{\epsilon}_1$ and $\bm{\epsilon}_2$ are unobserved errors and ${\bm \gamma}_0\in\mathcal{R}^{s_1 \times m} $, ${\bm \gamma}_1\in\mathcal{R}^{ s_1 \times p}$, ${\bm \gamma}_2\in\mathcal{R}^{m\times p}$, ${\bm \theta}_1\in\mathcal{R}^{m \times d}$, and ${\bm \theta}_2\in\mathcal{R}^{p\times d}$ are parameters. We can proceed as before by intervening on $Z$, that is, by setting $Z = z^\prime$. In this case, we want to cancel the remaining information from ${ D}$ to ${Y}$ by intervening on $Z$: Figure \ref{fig:unresolved} illustrates this procedure. The non-gender-specific word matrix $\widehat{{\mbf Y}}$ with unresolved bias removed is
\begin{equation}
\widehat{{\mbf Y}} = {\mbf Z}\widehat{\bm \theta}_1.  
\end{equation}
This debiasing procedure does not retain any information of $\mbf{Y}_{\Delta^\perp}$. Therefore we  restore the information from $\mbf{Y}_{\Delta^\perp}$ by taking a similar way to the previous procedure.


\setcounter{algorithm}{1}
\begin{algorithm}[htb]
\caption{(U-DeSIP) Removing unresolved bias.}
\label{alg:FramworkU}
\begin{algorithmic}[1] 
\renewcommand{\algorithmicrequire}{{{\bf Input}:}}
\REQUIRE 
$\mbf{D}$, 
$\mbf{Z}$, 
$\mbf{X}$, 
and $\mbf{Y}$.
\STATE Solve ${\mbf Y} =  {\mbf Z}{\bm \theta}_1 +  {\mbf X}{\bm \theta}_2$ by PLS to get ($\widehat{\bm \theta}_1$, $\widehat{\bm\theta}_2$)
\STATE Compute $\widehat{\mbf{Y}} ={\mbf Z}\widehat{\bm \theta}_1$
\STATE Compute $\widehat{\mbf{Y}}_{\Delta^\perp} = \mbf{Y} - \mbf{D} (\mbf{D}^T \mbf{D})^{-1} \mbf{D}^T \mbf{Y}$ 
\STATE Compute $\widehat{\mbf{Y}}_{\text{U-DeSIP}} = \widehat{\mbf{Y}} + \widehat{\mbf{Y}}_{\Delta^\perp}$
\renewcommand{\algorithmicrequire}{{{\bf Output}:}}
\REQUIRE
$\widehat{\mbf{Y}}_{\text{U-DeSIP}}$ as debiased word matrix. 
\end{algorithmic}
\end{algorithm}

\section{Experiments}
In this section, we compare the proposed methods against other debiasing algorithms in a set of comprehensive experiments. Our results show that the proposed methods not only reduce bias in various evaluation tasks, but also enhance the performance of word embeddings in semantic evaluation tasks. Our debiasing methods outperform in downstream part-of-speech (POS) tagging, POS chunking, and named-entity recognition tasks.

We apply the proposed debiasing methods to 300-dimensional GloVe embeddings pre-trained on English Wikipedia data with 322,636 unique words \citep{pennington2014glove}.
As baselines, we also compare our results against previous state-of-the-art debiasing methods, including the hard-debiasing method (Hard) \citep{genderpca}, the gender-preserving debiasing method (GP)  \citep{kaneko2019gender}, word vector learning method (GN) \citep{zhao2018learning}, and the half-sibling regression debiasing method (HSR) \citep{yang2020causal}.
For a fair comparison, we utilize the other authors' implementations. \footnote{\text{https://github.com/Lei-Ding07/Word\_Debias\_DeSIR}}

To separate the words in the following experiments, we manually pick $11$ pairs of pure gender words such as  (\emph{he}, \emph{she}) and (\emph{him}, \emph{her})\footnote{See the accompanying appendix for details of word list}. We form $\mbf{D}$ using the differences between the vector embeddings corresponding to these word pairs. We similarly compute $\mbf{P}$ using the gendered word pairs listed in \citet{zhao2018learning}.
The words represented in $\mbf{P}$ contain significant non-gender-related information and gender-related information, e.g., \emph{bride} and \emph{bridegroom}. We choose the 50,000 most frequent words in GloVe to form $\mbf{Y}$, which contains the words to be debiased, following the evaluation procedure in \citet{gonen}; $\mbf{X}$ is formed using the remaining words. In all of the below experiments, we use a fixed screening parameter $\gamma_n = 0.92$ in P-DeSIP and $\gamma_n = 0.80$ in U-DeSIP.

\subsection{Quantitative Evaluation for Bias Tasks}
Throughout this section, the top $N$ gender-biased words are chosen by evaluating dot products with the gender direction \wemb{he}$-$\wemb{she} in the original word embedding (i.e. GloVe) and choosing the most positive and negative values as the most male- and female-biased words, respectively.

\subsubsection{Bias-by-projection Task.}
Bias-by-projection uses the dot product between the gender direction \wemb{he}$-$\wemb{she} and the word to be tested.
We compute and average the absolute projection bias of the top 50,000 most frequent words.

The first column of Table \ref{tab:subsec:projection} shows that our methods achieve very good results. Its performance is just below that of Hard-GloVe, which can be explained by the fact that Hard-Glove is trained by removing projections along the gender direction.

\subsubsection{Sembias Analogy Task.}
The SemBias test was first introduced in \citet{zhao2018learning} as a set of word analogy tests.
The task is to find the word pair in best analogy to the pair (\emph{he}, \emph{she}) among four options: a gender-specific word pair, e.g., (\emph{waiter}, \emph{waitress}); a gender-stereotype word pair, e.g., (\emph{doctor}, \emph{nurse}); and two highly-similar, bias-free word pairs, e.g. (\emph{dog}, \emph{cat}). The dataset contains 440 instances, of which 40 instances, denoted by SemBias(subset), are not used during training. We report accuracy in identifying gender-specific word pairs. 


The second and third columns of Table \ref{tab:subsec:projection} quantify accuracy in identifying gender-specific word pairs. Our P-DeSIP methods achieve very good performance in both tasks. Specifically, in the subset test, P-DeSIP outperforms GloVe by almost 40\%.

\begin{table}[ht]
\fontsize{9}{10}\selectfont
\centering
\begin{tabular}{@{}l c c c @{}}
\toprule
 &
  \begin{tabular}[c]{@{}c@{}} Bias-by-projection \end{tabular} & 
  \begin{tabular}[c]{@{}c@{}} SemBias\end{tabular} & 
  \begin{tabular}[c]{@{}c@{}} SemBias (subset)\end{tabular}  \\ \midrule
  
GloVe              & 0.0375 &   0.8023 &  0.5750  \\
Hard        & \textbf{0.0007} &  0.8250 &  0.3250  \\
GP        & 0.0366 &  0.8432 &  0.6500    \\
GN                  & 0.0555 &  \textbf{0.9773} &  \underline{0.7500}  \\
HSR           & 0.0218 &  0.8591 &  0.1000    \\
\midrule
P-DeSIP                   & \underline{0.0038} &  \underline{0.9523} &  \textbf{0.9750}  \\
U-DeSIP                   & \underline{0.0038} &  0.9090 &  0.5000  \\
\bottomrule
\end{tabular}
\caption{Gender-direction-related task performance. In each column, the best and second-best results are boldfaced and underlined, respectively.}
\label{tab:subsec:projection}
\end{table}


\subsubsection{Clustering Male- and Female-biased Words.}
As noted in \citet{gonen}, biased words tend to cluster together. Even some debiased embeddings were unable to escape from this phenomenon. Here we take the top 500 male-biased words and the top 500 female-biased words and partition them via K-means clustering (K=2) \citep{hartigan1979algorithm}. 
Accuracy in splitting the 1,000 words into male and female clusters is presented in Table \ref{tab:subsec:GBWR}. Our methods achieve the best performance among all other methods.

\subsubsection{Correlation between Bias-by-projection and Bias-by Neighbors.}
Taking again the top 50,000 most frequent words as targets, we compute the Pearson correlation coefficient between the bias-by-projection and bias-by-neighbor results.
The latter is computed using the neighborhood metric, which counts the percentage of male- and female-biased words within the $K$-nearest neighbors of each target word \citep{gonen, wang2020double}. Here, we take $K=100$.
Referring to the second column of Table \ref{tab:subsec:GBWR}, our methods generally achieve the best performance.

\subsubsection{Bias-by-neighbors for Profession Words.}
In this task, we assess the effect of debiasing by calculating the correlation between bias-by-neighbor measures before and after debiasing.
We use the neighborhood metric, as in the previous task, but we restrict our targets to the list of professional words in \citet{genderpca} and \citet{zhao2018learning}.
Results, in the third column of Table \ref{tab:subsec:GBWR}, show that our methods outperform GloVe and are comparable to HSR-GloVe.

\subsubsection{Classifying Previously Female- and Male-biased Words.}
After selecting the top 2,500 biased words for each gender, for each baseline model we train a support vector machine (SVM) model using 1,000 randomly sampled words.
This classifier is then applied to the remaining 4,000 words to predict gender bias direction.
Prediction accuracy is shown in the last column of Table \ref{tab:subsec:GBWR}: a lower accuracy indicates the trained model is unable to capture gender-related information from the original embedding and thus, that the debiasing method is superior.
Again, both of our methods outperform the other methods.

\begin{table}[ht]
\fontsize{9}{10}\selectfont
\centering
\begin{tabular}{@{}lcccc@{}}
\toprule
 &
  \begin{tabular}[c]{@{}c@{}} Clustering \end{tabular} &
  \begin{tabular}[c]{@{}c@{}} Correlation\end{tabular} &
  \begin{tabular}[c]{@{}c@{}} Profession\end{tabular} &
  \begin{tabular}[c]{@{}c@{}} Classification\end{tabular}
  \\ \midrule
  
GloVe              & 1.0000 &  0.7727 & 0.8200 & 0.9980 \\
Hard        &  0.8050 & 0.6884 & 0.7161 & 0.9068  \\
GP                    & 1.0000 & 0.7700 & 0.8102 & 0.9978    \\
GN                    & 0.8560 & 0.7336 & 0.7925 & 0.9815    \\
HSR                    & 0.9410  & \underline{0.6422} & \textbf{0.6804} & 0.9055   \\
\midrule
P-DeSIP   & \textbf{0.7910} &  0.6431 &  0.7096 & \textbf{0.8547}  \\
U-DeSIP   & \underline{0.7920} & \textbf{0.6421} &  \underline{0.7060} &\underline{0.8550} \\ 
\bottomrule
\end{tabular}
\caption{Gender bias word relation task performance. In each column, the best and second-best results are boldfaced and underlined, respectively.}
\label{tab:subsec:GBWR}
\end{table}

\subsubsection{Word Embedding Association Test (WEAT)}

The WEAT test \citep{weatscience} is a permutation-based test that measures bias in word embeddings. We report effect sizes ($d$) and p-values ($p$) in our results. The effect size is a normalized measure of how separated two distributions are.  A higher value indicates a larger bias between target words with respect to attribute words. The $p$-values denote whether the bias is significant or not.

We conduct three tests using the Pleasant \& Unpleasant (Task 1), Career \& Family (Task 2), and Science \& Art (Task 3) word sets. We consider male and female names as attribute sets.\footnote{All word lists are from \citet{weatscience}. Because GloVe embeddings are uncased, we use lower case words.}. As shown in Table \ref{tab:subsec:WEAT}, we achieve results comparable to those for other methods. In two out of three tasks, the $p$-value is not significant. We also achieve a reasonably small effect size in all three tasks.

\begin{table}
\fontsize{9}{10}\selectfont
  \begin{tabular}{p{1.2cm}p{0.63cm}cp{0.63cm}cp{0.63cm}c}
    \toprule
    \multirow{2}{*}{} &
      \multicolumn{2}{c}{Task1} &
      \multicolumn{2}{c}{Task2} &
      \multicolumn{2}{c}{Task3}  \\
      \cmidrule(lr){2-3} \cmidrule(lr){4-5} \cmidrule(lr){6-7} 
      & $p$ & $d$ & $p$ & $d$ & $p$ & $d$ \\
    \midrule
    
    GloVe & 0.090$^*$ & 0.704 & 0.000 & 1.905 & 0.026 & 0.987 \\
    
    Hard & 0.363$^*$ & \textbf{0.187} & 0.000 & 1.688 & 0.583$^*$ & -0.104 \\
    GP & 0.055$^*$ & 0.832 & 0.000 & 1.909 & 0.025 & 0.997 \\
    GN & 0.157$^*$ & 0.541 & 0.074$^*$ & \textbf{0.753} & 0.653$^*$ & -0.222 \\
    
    HSR & 0.265$^*$ & 0.340 & 0.000 & 1.555 & 0.410$^*$ & 0.122 \\
    \midrule
    P-DeSIP & 0.755$^*$ & -0.373 & 0.001 & \underline{1.459} & 0.486$^*$ & \underline{0.019} \\
    U-DeSIP & 0.732$^*$ & \underline{-0.335} & 0.001 & 1.462 & 0.491$^*$ & \textbf{0.012} \\
    \bottomrule
  \end{tabular}
  \caption{WEAT test result. In each column of $p$-value, $^*$  indicates statistically \textbf{non}-significant compare with $\alpha=0.05$; In each column of $d$, the  best  and  second-best  results  are  boldfaced  and  underlined, respectively.}
\label{tab:subsec:WEAT}
\end{table}

\subsection{Visualization}
\begin{figure}[!htbp]
\captionsetup[subfigure]{justification=centering}
\centering
 \subcaptionbox{GloVe}[0.45\linewidth]
 {\includegraphics[scale = 0.33]{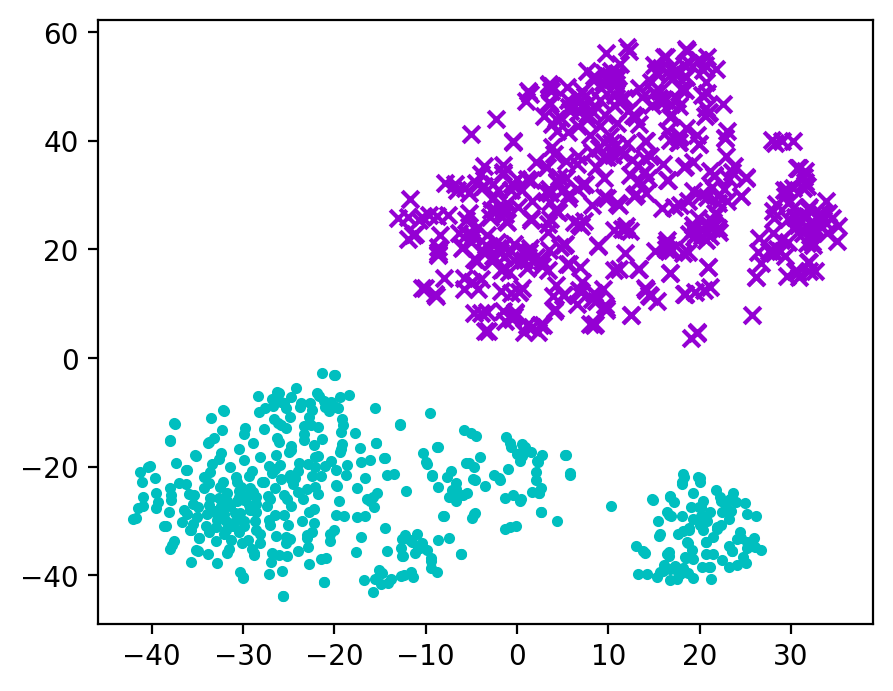}}
 \subcaptionbox{Hard-debias}[0.45\linewidth]
 {\includegraphics[scale = 0.33]{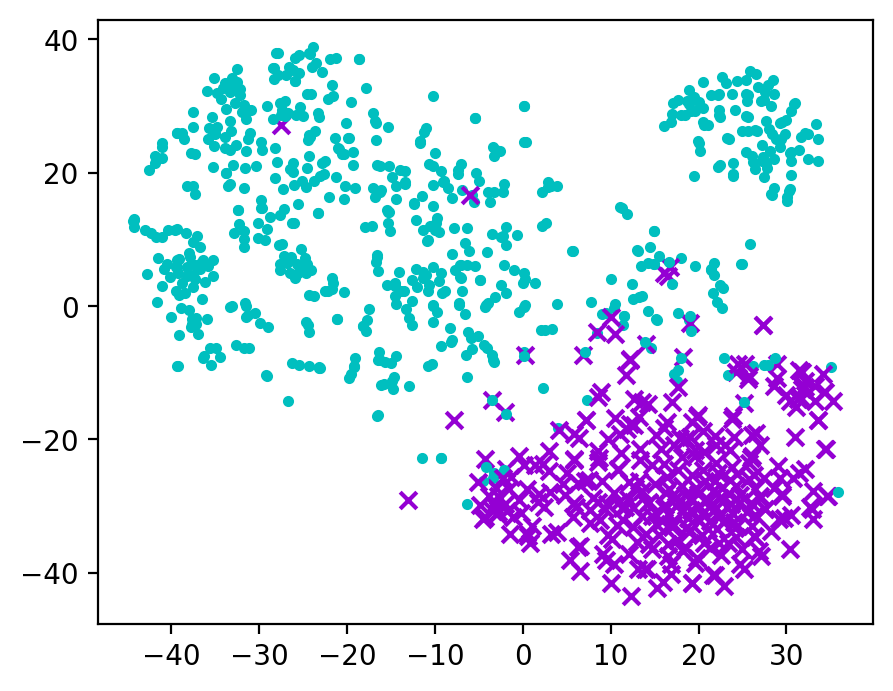}}
 \subcaptionbox{GP-debias}[0.45\linewidth]
 {\includegraphics[scale = 0.33]{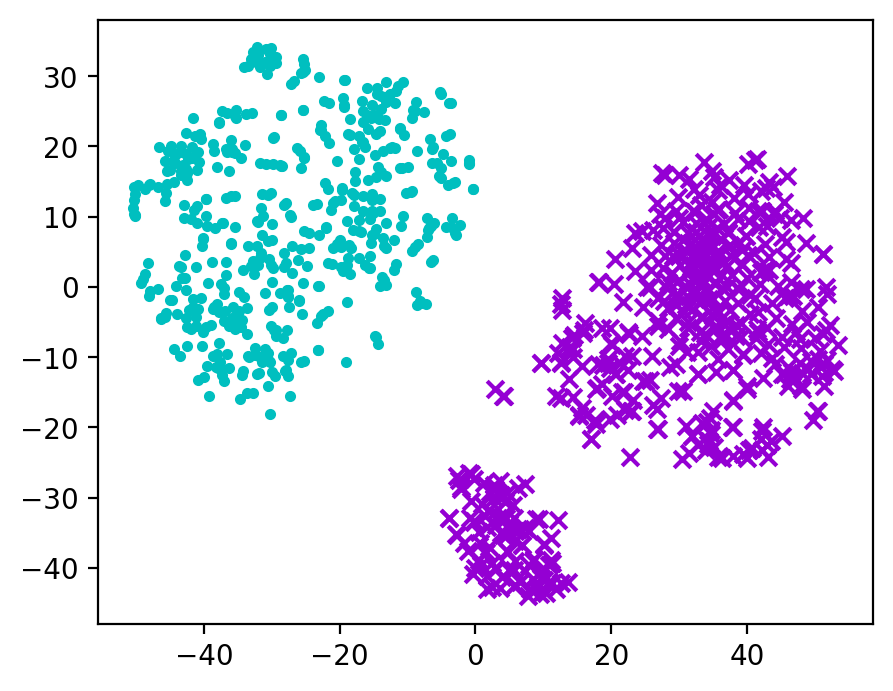}}
 \subcaptionbox{HSR}[0.45\linewidth]
 {\includegraphics[scale = 0.33]{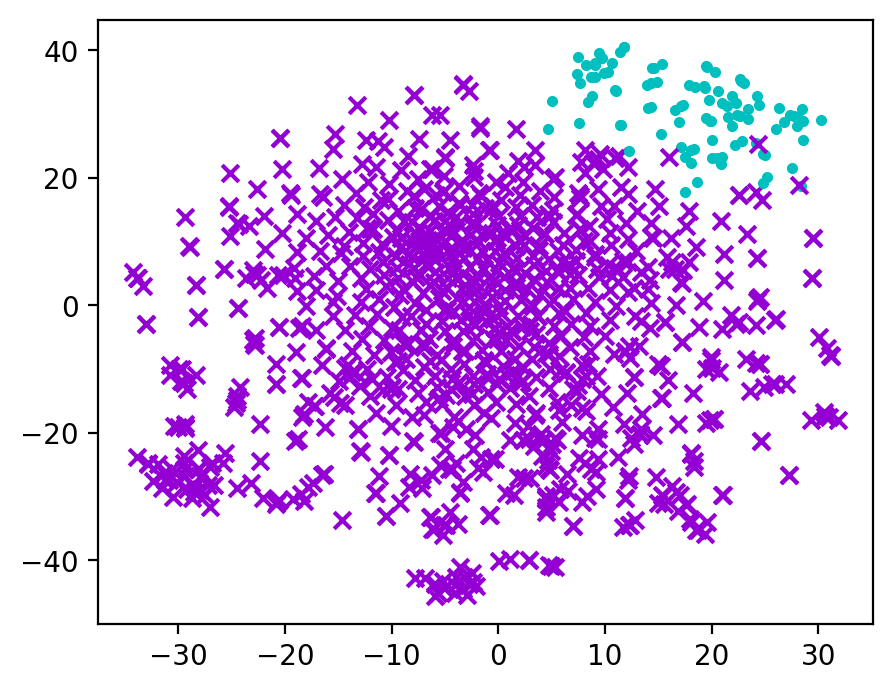}}

 \subcaptionbox{P-DeSIP}[0.45\linewidth]
 {\includegraphics[scale = 0.33]{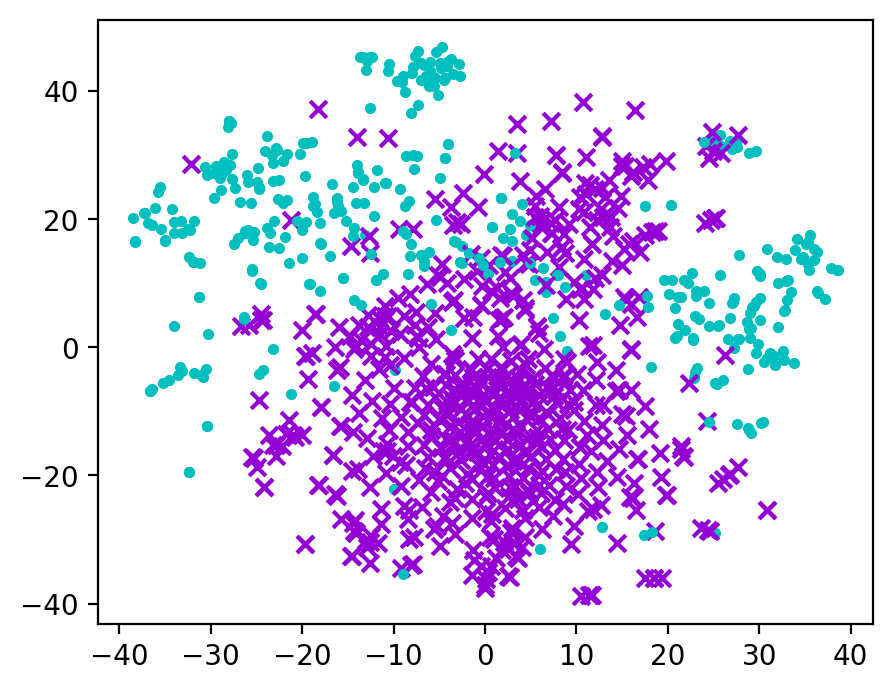}}
 \subcaptionbox{U-DeSIP}[0.45\linewidth]
 {\includegraphics[scale = 0.33]{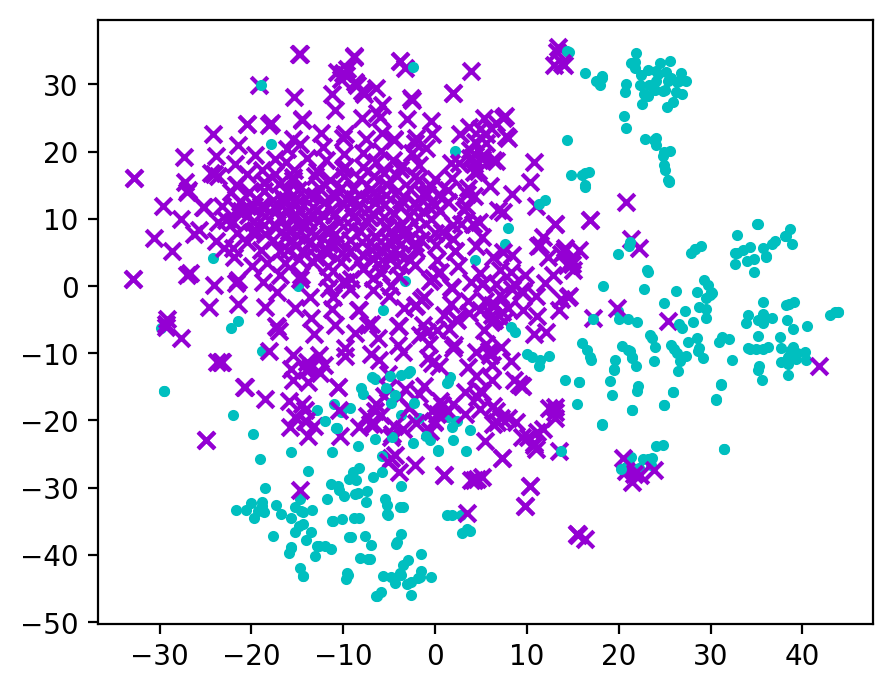}}

\caption{t-SNE visualization. }
\label{fig:tsne}
\end{figure}

In order to visually illustrate that our proposed methods effectively reduce gender bias, we took the top 500 male- and female-biased embeddings and generated a t-SNE projection \citep{hinton2002stochastic} for all of the baseline embeddings. In Figure \ref{fig:tsne}, the two colors in the graphs indicate male- and female-biased embeddings.  We can see our two methods more effectively mix up the male- and female-biased embeddings.

\begin{table*}[!t]
\centering
\fontsize{9}{10}\selectfont
  \begin{tabular}{lccccccccc}
    \toprule
    \multicolumn{10}{c}{Embedding Matrix Replacement} \\
    \midrule
    \multirow{2}{*}{} &
      \multicolumn{3}{c}{POS Tagging} &
      \multicolumn{3}{c}{POS Chunking} &
      \multicolumn{3}{c}{Named Entity Recognition} \\
      \cmidrule(lr){2-4} \cmidrule(lr){5-7} \cmidrule(lr){8-10}
       & $\Delta$ \rm F1 & $\Delta$ \rm Precision & $\Delta$ \rm Recall & $\Delta$ \rm F1 & $\Delta$ \rm Precision & $\Delta$ \rm Recall & $\Delta$ \rm F1 & $\Delta$ \rm Precision & $\Delta$ \rm Recall \\
    \midrule

    Hard & -0.0776 & -0.0736 & -0.2079 & -0.0653 & -0.1500 & -0.1009 & -0.0118 & -0.0187 & -0.0238 \\
    GP & -0.1021 & -0.1910 & -0.2068 & -0.0702 & -0.1385 & -0.1301 & -0.0353 & -0.0366 & -0.0871\\
    GN & -0.0987 & -0.1001 & -0.2554 & -0.0702 & -0.1269 & -0.1401 & -0.0294 & -0.0610 & -0.0472\\
    HSR & -0.0666 & -0.0589 & -0.1820 &  -0.0377 & -0.0753 & -0.0689 & -0.0055 & -0.0068 & -0.0128 \\
    \midrule
    P-DeSIP & \underline{-0.0133} & \underline{-0.0006} & \underline{-0.0471} & \textbf{-0.0108} & \textbf{-0.0036} & \underline{-0.0346} & \underline{-0.0014} & \textbf{0.0002} & \underline{-0.0052} \\
    U-DeSIP  & \textbf{-0.0107} & \textbf{0.0033} & \textbf{-0.0405} & \underline{-0.0110} & \underline{-0.0073} & \textbf{-0.0324} & \textbf{-0.0007} & \underline{0.0013} & \textbf{-0.0035} \\
    \midrule
    \multicolumn{10}{c}{Model Retraining} \\
    \midrule
    \multirow{2}{*}{} &
      \multicolumn{3}{c}{POS Tagging} &
      \multicolumn{3}{c}{POS Chunking} &
      \multicolumn{3}{c}{Named Entity Recognition} \\
      \cmidrule(lr){2-4} \cmidrule(lr){5-7} \cmidrule(lr){8-10}
       & $\Delta$ \rm F1 & $\Delta$ \rm Precision & $\Delta$ \rm Recall & $\Delta$ \rm F1 & $\Delta$ \rm Precision & $\Delta$ \rm Recall & $\Delta$ \rm F1 & $\Delta$ \rm Precision & $\Delta$ \rm Recall \\
    \midrule

    Hard & -0.0194 & \underline{0.0078} & -0.0741 & -0.0106 & \textbf{0.0075} & -0.0438 & -0.0050 & \textbf{0.0013} & -0.0179 \\
    GP & -0.0071 & 0.0011 & -0.0264 & -0.0069 & \underline{0.0043} & -0.0278 & -0.0013 & -0.0014 & -0.0030\\
    GN & -0.0027 & \textbf{0.0089} & -0.0174 & 0.0000 & -0.0074 & 0.0067 & -0.0011 & -0.0254 & \textbf{0.0189}\\
    HSR & -0.0055 & -0.0009 & -0.0192 & \underline{0.0002} & -0.0089 & \underline{0.0084} & -0.0017 & -0.0011 & -0.0050 \\
    \midrule
     P-DeSIP & \underline{-0.0018} & 0.0002 & \underline{-0.0068} & -0.0005 & -0.0041 & 0.0016 & \underline{0.0002} & -0.0007 & 0.0011 \\
      U-DeSIP  & \textbf{-0.0010} & 0.0000 & \textbf{-0.0036} & \textbf{0.0032} & -0.0009 & \textbf{0.0125} & \textbf{0.0005} & \underline{0.0008} & \underline{0.0013} \\
    \bottomrule
  \end{tabular}
  \centering
  \caption{Result of downstream tasks, positive value means the task has better performance than using Original GloVe. In each column, the best and second-best results are boldfaced and underlined, respectively.}
\label{tab:subsec:downstream}
\end{table*}

\subsection{Word Similarity Tasks}
Another important aspect of word embedding is its ability to encode words' semantic information.
While bias removal is our main goal, it is unacceptable to disregard how semantic information is influenced by the debiasing process.
We next implement several word similarity tests to evaluate our algorithms against existing baseline methods.
We consider the following tasks: RG65 \citep{rg65}, WordSim-353 \citep{wordsim353}, Rarewords \citep{rarewords}, MEN \citep{men}, MTurk-287 \citep{mturk287}, and MTurk-771 \citep{mturk771}. \emph{SimLex-999} \citep{simlex999}, and \emph{SimVerb-3500} \citep{simverb3500}.
These datasets associated with each task contain word pairs and a corresponding human-annotated similarity score.

As an evaluation measure, we compute Spearman's rank correlation coefficient between these two ranks. Results are shown in Table \ref{tab:subsec:wordsim1} and \ref{tab:subsec:wordsim2}.  We see that our methods have the leading performance for most of the tasks.

\begin{table}[ht]
\fontsize{9}{10}\selectfont
\centering
\begin{tabular}{@{}l rrrr}
\toprule &

  \begin{tabular}[c]{@{}c@{}} RG65 \end{tabular} &
  \begin{tabular}[c]{@{}c@{}} WS\end{tabular} &
  \begin{tabular}[c]{@{}c@{}} RW\end{tabular} &
  \begin{tabular}[c]{@{}c@{}} MEN\end{tabular} \\
   \midrule
  
GloVe             & 0.7540 & 0.6199 & 0.3722 & 0.7216 \\
Hard        & 0.7648 & 0.6207 & 0.3720 & 0.7212 \\
GP          & 0.7546 & 0.6003 & 0.3450 & 0.6974   \\
GN          & 0.7457 & 0.6286 & \textbf{0.3989} & 0.7446  \\
HSR         & 0.7764 & 0.6554 & 0.3868 & 0.7353   \\
\midrule
P-DeSIP   & \textbf{0.7794} &  \textbf{0.6856} &  \underline{0.3970} & \textbf{0.7484} \\
U-DeSIP   & \underline{0.7734} & \underline{0.6828} &  0.3956 &\underline{0.7478} \\ 
\bottomrule
\end{tabular}
\caption{Word similarity task performance 1. In each column, the best and second-best results are boldfaced and underlined, respectively.}
\label{tab:subsec:wordsim1}
\end{table}

\begin{table}[ht]
\fontsize{9}{10}\selectfont
\centering
\begin{tabular}{@{}l rrrr}
\toprule &

  \begin{tabular}[c]{@{}c@{}} MT-287\end{tabular} &
  \begin{tabular}[c]{@{}c@{}} MT-771\end{tabular} &
  \begin{tabular}[c]{@{}c@{}} SimLex\end{tabular} &
  \begin{tabular}[c]{@{}c@{}} SimVerb\end{tabular}
  \\ \midrule
  
GloVe       & \underline{0.6480} & 0.6486 & 0.3474 & 0.2038\\
Hard         & 0.6468 & 0.6504 & 0.3501 & 0.2034\\
GP           & 0.6418 & 0.6391 & 0.3389 & 0.1877 \\
GN           & \textbf{0.6617} & 0.6619 & 0.3700 & 0.2219 \\
HSR         & 0.6335 & 0.6652  & \textbf{0.3971} & \textbf{0.2635} \\
\midrule
P-DeSIP    & 0.6452 &  \textbf{0.6741} & \underline{0.3765} & \underline{0.2286}\\
U-DeSIP    & 0.6455 & \underline{0.6731} & 0.3756 & 0.2273\\ 
\bottomrule
\end{tabular}
\caption{Word similarity task performance 2. In each column, the best and second-best results are boldfaced and underlined, respectively.}
\label{tab:subsec:wordsim2}
\end{table}

\subsection{Downstream Task Utility Evaluation}
In order to demonstrate that our de-biased word embeddings still retain good downstream utility and performance, we follow the CoNLL2003 shared task \citep{sang2003introduction} and use POS tagging, POS chunking, and named-entity recognition(NER) as the evaluation tasks. Following \citet{multiclass} we evaluate each task in two ways: embedding matrix replacement and model retraining. 

In embedding matrix replacement, we first train the task model using the original biased GloVe vectors and then calculate test data performance differences when using the original biased GloVe embeddings versus other debiased embeddings. 
Table \ref{tab:subsec:downstream} suggests constant performance degradation for all debiasing methods relative to the original embedding. Despite this, our methods outperform all the other tasks (in the sense of minimizing degradation) by a large margin across all the tasks and evaluation metrics (i.e., F1 score, precision, and recall).
Furthermore, we even achieve a small improvement in precision on the NER task.

In model retraining, we first train two task models, one using the original biased GloVe embeddings and the other using debiased embeddings. We then calculate differences in test performance.
Table \ref{tab:subsec:downstream} again suggests that our methods have the closest performance to the model trained and tested using the original GloVe embeddings. Our method also displays the most consistent and comparable performance across the three tasks.

\section{Conclusion}
In this paper, we develop two causal inference methods for removing biases in word embeddings. We show that using  the  differences  between vectors corresponding to paired gender-specific words can better represent and eliminate gender bias. 
We find an intuitive and effective way to better represent gender information that needs to be removed and use this approach to achieve oracle-like retention of semantic and lexical information. We also show that our methods outperform other debiasing methods in downstream NLP tasks. Furthermore, our methods easily accommodate situations where other kinds of bias exist, such as social, racial, or class biases.

There are several important directions for future work. 
First, we only consider the linear relationship among the proposed casual inference frameworks. Further investigation is warranted to extend these frameworks to incorporate the non-linear causal relationship \cite{hoyer2008nonlinear}. 
Second, when $P$ is not attainable, we select the resolving variables $Z$ to contain the adjectives and nouns correlated to gender bias variables $D$. This selection method is rather heuristic. If prior knowledge about resolving variables was introduced, it would surely improve the performance of the unresolved bias removal.
Third, we introduce a residual block to restore the information not retained from the debiasing procedure. The construction of it is rather intuitive and requires more rigorous justification. 
Fourth, incorporating other dimension reduction techniques such as wavelet and spline methods \citep{yu2019sparse} are deemed for further explorations. 
Finally, although our methods facilitate easy accommodations for situations where other kinds of bias exist, how the proxy and resolving variables as well as the bias variables are properly pre-specified may require non trivial efforts.


\appendix
\newpage
\section{Acknowledgments}
This work was supported by the Economic and Social Research Council (ESRC ES/T012382/1) and the Social Sciences and Humanities Research Council (SSHRC 2003-2019-0003)  under the scheme of the Canada-UK Artificial Intelligence Initiative. The project title is BIAS: Responsible AI for Gender and Ethnic Labour Market Equality.
We appreciated all the constructive comments from the reviewers. We thank Ketong Shen for his helpful discussion and valuable input.
\bibliography{main}

\begin{thebibliography}{35}
\providecommand{\natexlab}[1]{#1}

\bibitem[{Bansal et~al.(2021)Bansal, Garimella, Suhane, and
  Mukherjee}]{bansal2021debiasing}
Bansal, S.; Garimella, V.; Suhane, A.; and Mukherjee, A. 2021.
\newblock Debiasing Multilingual Word Embeddings: A Case Study of Three Indian
  Languages.
\newblock In \emph{Proceedings of the 32nd ACM Conference on Hypertext and
  Social Media}, 27--34.

\bibitem[{Bolukbasi et~al.(2016)Bolukbasi, Chang, Zou, Saligrama, and
  Kalai}]{genderpca}
Bolukbasi, T.; Chang, K.-W.; Zou, J.~Y.; Saligrama, V.; and Kalai, A.~T. 2016.
\newblock Man is to computer programmer as woman is to homemaker? debiasing
  word embeddings.
\newblock \emph{Advances in Neural Information Processing Systems}, 29:
  4349--4357.

\bibitem[{Bruni, Tran, and Baroni(2014)}]{men}
Bruni, E.; Tran, N.-K.; and Baroni, M. 2014.
\newblock Multimodal distributional semantics.
\newblock \emph{Journal of artificial intelligence research}, 49: 1--47.

\bibitem[{Caliskan, Bryson, and Narayanan(2017)}]{weatscience}
Caliskan, A.; Bryson, J.~J.; and Narayanan, A. 2017.
\newblock Semantics derived automatically from language corpora contain
  human-like biases.
\newblock \emph{Science}, 356(6334): 183--186.

\bibitem[{Dev and Phillips(2019)}]{dev2019attenuating}
Dev, S.; and Phillips, J. 2019.
\newblock Attenuating bias in word vectors.
\newblock In \emph{The 22nd International Conference on Artificial Intelligence
  and Statistics}, 879--887. PMLR.

\bibitem[{Douglas(2017)}]{douglas2017ai}
Douglas, L. 2017.
\newblock AI is not just learning our biases; it is amplifying them.
\newblock \emph{Medium, December}, 5.

\bibitem[{Fan and Lv(2008)}]{fan2008sure}
Fan, J.; and Lv, J. 2008.
\newblock Sure independence screening for ultrahigh dimensional feature space.
\newblock \emph{Journal of the Royal Statistical Society: Series B (Statistical
  Methodology)}, 70(5): 849--911.

\bibitem[{Finkelstein et~al.(2001)Finkelstein, Gabrilovich, Matias, Rivlin,
  Solan, Wolfman, and Ruppin}]{wordsim353}
Finkelstein, L.; Gabrilovich, E.; Matias, Y.; Rivlin, E.; Solan, Z.; Wolfman,
  G.; and Ruppin, E. 2001.
\newblock Placing search in context: The concept revisited.
\newblock In \emph{Proceedings of the 10th international conference on World
  Wide Web}, 406--414.

\bibitem[{Garg et~al.(2018)Garg, Schiebinger, Jurafsky, and Zou}]{rnd}
Garg, N.; Schiebinger, L.; Jurafsky, D.; and Zou, J. 2018.
\newblock Word embeddings quantify 100 years of gender and ethnic stereotypes.
\newblock \emph{Proceedings of the National Academy of Sciences}, 115(16):
  E3635--E3644.

\bibitem[{Gerz et~al.(2016)Gerz, Vuli{\'c}, Hill, Reichart, and
  Korhonen}]{simverb3500}
Gerz, D.; Vuli{\'c}, I.; Hill, F.; Reichart, R.; and Korhonen, A. 2016.
\newblock Simverb-3500: A large-scale evaluation set of verb similarity.
\newblock \emph{arXiv preprint arXiv:1608.00869}.

\bibitem[{Gonen and Goldberg(2019)}]{gonen}
Gonen, H.; and Goldberg, Y. 2019.
\newblock Lipstick on a Pig: Debiasing Methods Cover up Systematic Gender
  Biases in Word Embeddings But do not Remove Them.
\newblock \emph{NAACL-HLT}.

\bibitem[{Greenwald, McGhee, and Schwartz(1998)}]{greenwald1998measuring}
Greenwald, A.~G.; McGhee, D.~E.; and Schwartz, J.~L. 1998.
\newblock Measuring individual differences in implicit cognition: the implicit
  association test.
\newblock \emph{Journal of personality and social psychology}, 74(6): 1464.

\bibitem[{Halawi et~al.(2012)Halawi, Dror, Gabrilovich, and Koren}]{mturk771}
Halawi, G.; Dror, G.; Gabrilovich, E.; and Koren, Y. 2012.
\newblock Large-scale learning of word relatedness with constraints.
\newblock In \emph{Proceedings of the 18th ACM SIGKDD international conference
  on Knowledge discovery and data mining}, 1406--1414.

\bibitem[{Hartigan and Wong(1979)}]{hartigan1979algorithm}
Hartigan, J.~A.; and Wong, M.~A. 1979.
\newblock Algorithm AS 136: A k-means clustering algorithm.
\newblock \emph{Journal of the royal statistical society. series c (applied
  statistics)}, 28(1): 100--108.

\bibitem[{Hill, Reichart, and Korhonen(2015)}]{simlex999}
Hill, F.; Reichart, R.; and Korhonen, A. 2015.
\newblock Simlex-999: Evaluating semantic models with (genuine) similarity
  estimation.
\newblock \emph{Computational Linguistics}, 41(4): 665--695.

\bibitem[{Hinton and Roweis(2002)}]{hinton2002stochastic}
Hinton, G.; and Roweis, S.~T. 2002.
\newblock Stochastic neighbor embedding.
\newblock In \emph{NIPS}, volume~15, 833--840. Citeseer.

\bibitem[{Hoyer et~al.(2008)Hoyer, Janzing, Mooij, Peters, Sch{\"o}lkopf
  et~al.}]{hoyer2008nonlinear}
Hoyer, P.~O.; Janzing, D.; Mooij, J.~M.; Peters, J.; Sch{\"o}lkopf, B.; et~al.
  2008.
\newblock Nonlinear causal discovery with additive noise models.
\newblock In \emph{NIPS}, volume~21, 689--696. Citeseer.

\bibitem[{Kaneko and Bollegala(2019)}]{kaneko2019gender}
Kaneko, M.; and Bollegala, D. 2019.
\newblock Gender-preserving debiasing for pre-trained word embeddings.
\newblock \emph{arXiv preprint arXiv:1906.00742}.

\bibitem[{Kilbertus et~al.(2017)Kilbertus, Rojas-Carulla, Parascandolo, Hardt,
  Janzing, and Sch{\"o}lkopf}]{kilbertus2017avoiding}
Kilbertus, N.; Rojas-Carulla, M.; Parascandolo, G.; Hardt, M.; Janzing, D.; and
  Sch{\"o}lkopf, B. 2017.
\newblock Avoiding discrimination through causal reasoning.
\newblock \emph{arXiv preprint arXiv:1706.02744}.

\bibitem[{Luong, Socher, and Manning(2013)}]{rarewords}
Luong, M.-T.; Socher, R.; and Manning, C.~D. 2013.
\newblock Better word representations with recursive neural networks for
  morphology.
\newblock In \emph{Proceedings of the seventeenth conference on computational
  natural language learning}, 104--113.

\bibitem[{Manzini et~al.(2019)Manzini, Lim, Tsvetkov, and Black}]{multiclass}
Manzini, T.; Lim, Y.~C.; Tsvetkov, Y.; and Black, A.~W. 2019.
\newblock Black is to criminal as caucasian is to police: Detecting and
  removing multiclass bias in word embeddings.
\newblock \emph{NAACL}.

\bibitem[{Mikolov et~al.(2013)Mikolov, Chen, Corrado, and
  Dean}]{mikolov2013efficient}
Mikolov, T.; Chen, K.; Corrado, G.; and Dean, J. 2013.
\newblock Efficient estimation of word representations in vector space.
\newblock \emph{arXiv preprint arXiv:1301.3781}.

\bibitem[{Pennington, Socher, and Manning(2014)}]{pennington2014glove}
Pennington, J.; Socher, R.; and Manning, C.~D. 2014.
\newblock Glove: Global vectors for word representation.
\newblock In \emph{Proceedings of the 2014 conference on empirical methods in
  natural language processing (EMNLP)}, 1532--1543.

\bibitem[{Radinsky et~al.(2011)Radinsky, Agichtein, Gabrilovich, and
  Markovitch}]{mturk287}
Radinsky, K.; Agichtein, E.; Gabrilovich, E.; and Markovitch, S. 2011.
\newblock A word at a time: computing word relatedness using temporal semantic
  analysis.
\newblock In \emph{Proceedings of the 20th international conference on World
  wide web}, 337--346.

\bibitem[{Rubenstein and Goodenough(1965)}]{rg65}
Rubenstein, H.; and Goodenough, J.~B. 1965.
\newblock Contextual correlates of synonymy.
\newblock \emph{Communications of the ACM}, 8(10): 627--633.

\bibitem[{Sang and De~Meulder(2003)}]{sang2003introduction}
Sang, E.~F.; and De~Meulder, F. 2003.
\newblock Introduction to the CoNLL-2003 shared task: Language-independent
  named entity recognition.
\newblock \emph{arXiv preprint cs/0306050}.

\bibitem[{Shin et~al.(2020)Shin, Song, Jang, Kim, Joo, and
  Moon}]{shin2020neutralizing}
Shin, S.; Song, K.; Jang, J.; Kim, H.; Joo, W.; and Moon, I.-C. 2020.
\newblock Neutralizing gender bias in word embedding with latent
  disentanglement and counterfactual generation.
\newblock \emph{arXiv preprint arXiv:2004.03133}.

\bibitem[{Vinzi et~al.(2010)Vinzi, Chin, Henseler, and
  Wang}]{vinzi2012handbook}
Vinzi, V.~E.; Chin, W.~W.; Henseler, J.; and Wang, H. 2010.
\newblock Handbook of Partial Least Squares: Concepts, Methods and
  Applications.
\newblock \emph{Springer}.

\bibitem[{Wang et~al.(2020)Wang, Lin, Rajani, McCann, Ordonez, and
  Xiong}]{wang2020double}
Wang, T.; Lin, X.~V.; Rajani, N.~F.; McCann, B.; Ordonez, V.; and Xiong, C.
  2020.
\newblock Double-hard debias: Tailoring word embeddings for gender bias
  mitigation.
\newblock \emph{arXiv preprint arXiv:2005.00965}.

\bibitem[{Xie et~al.(2020)Xie, Lin, Yan, and Tang}]{xie2020category}
Xie, J.; Lin, Y.; Yan, X.; and Tang, N. 2020.
\newblock Category-adaptive variable screening for ultra-high dimensional
  heterogeneous categorical data.
\newblock \emph{Journal of the American Statistical Association}, 115(530):
  747--760.

\bibitem[{Yang and Feng(2020)}]{yang2020causal}
Yang, Z.; and Feng, J. 2020.
\newblock A causal inference method for reducing gender bias in word embedding
  relations.
\newblock In \emph{Proceedings of the AAAI Conference on Artificial
  Intelligence}, volume~34, 9434--9441.

\bibitem[{Yu, Kong, and Mizera(2016)}]{yu2016partial}
Yu, D.; Kong, L.; and Mizera, I. 2016.
\newblock Partial functional linear quantile regression for neuroimaging data
  analysis.
\newblock \emph{Neurocomputing}, 195: 74--87.

\bibitem[{Yu et~al.(2019)Yu, Zhang, Mizera, Jiang, and Kong}]{yu2019sparse}
Yu, D.; Zhang, L.; Mizera, I.; Jiang, B.; and Kong, L. 2019.
\newblock Sparse wavelet estimation in quantile regression with multiple
  functional predictors.
\newblock \emph{Computational Statistics \& Data Analysis}, 136: 12--29.

\bibitem[{Zhao et~al.(2019)Zhao, Wang, Yatskar, Cotterell, Ordonez, and
  Chang}]{zhao2019gender}
Zhao, J.; Wang, T.; Yatskar, M.; Cotterell, R.; Ordonez, V.; and Chang, K.-W.
  2019.
\newblock Gender bias in contextualized word embeddings.
\newblock \emph{arXiv preprint arXiv:1904.03310}.

\bibitem[{Zhao et~al.(2018)Zhao, Zhou, Li, Wang, and Chang}]{zhao2018learning}
Zhao, J.; Zhou, Y.; Li, Z.; Wang, W.; and Chang, K.-W. 2018.
\newblock Learning gender-neutral word embeddings.
\newblock \emph{arXiv preprint arXiv:1809.01496}.

\end{thebibliography}

\section{Appendix}
\subsection{Detail explanation of Table 1}
For each of the four pre-determined words \emph{Wedding}, \emph{Service}, \emph{Family}, and \emph{Religion}, we identify the top $200$ most cosine-correlated words. For each of the $200$ words, we fit a ridge regression against gender-specific words defined in \citet{yang2020causal} (HSR), and a linear regression against the differences between gender-specific word pairs from this paper (DeSIP). The fitted word vectors are used as reduced-bias word vectors. To quantify the semantic information preservation, the mean absolute dot product between the pre-determined words and their bias-reduced versions over the $200$ most related words are presented, with standard errors in parentheses. Note that, the oracle preservation semantic information is achieved by using the original word vector instead of the fitted one. The last row shows the proportion of these $200$ words for which DeSIP outperforms HSR with respect to semantic information preservation.

\subsection{Pure gender word list of $\mbf{D}$}

\textit{Male words}: he, him, man, his, himself, son, father, guy, boy, male, men, sons, fathers, guys, boys, males, sir, gentleman, gentlemen, mr
\\
\textit{Female words}: she, her, woman, hers, herself, daughter, mother, gal, girl, female, women, daughters, mothers, gals, girls, females, madam, lady, ladies, mrs\\
$\mbf{D}$ is formed by subtraction of each word in Male words with the corresponding word in Female words.

\subsection{Detail derivation of equation (2) and (4)}
We present the details about how to obtain the equations (2) and (4) here as follows:
\begin{itemize}
    \item Intervene on $\mbf{P}$ by removing all incoming arrows, see Figure \ref{fig:proxy}, and set $\mbf{P}=p'$, where $p'$ is a random variable. Then we obtain: 
 \begin{align*}
\mbf{P} = p', 
\mbf{X} = \mbf{D} \bm\alpha_1 + \mbf{P} \bm\alpha_2  + \bm{e}_2,
\mbf{Y} = \mbf{P}\bm\beta_1 + \mbf{X}\bm\beta_2.    \end{align*}   
\item Integrate the first and second equations into the third equation from their structural equations. 
\begin{align*}
\mbf{Y} = p'(\bm\beta_1 + \bm\alpha_2\bm\beta_2) + (\mbf{D} \bm\alpha_1 + \bm{e}_2)\bm\beta_2.
\end{align*}
\item Require the distribution of $\mbf{Y}$ to be independent of $p'$, i.e. for all $p_1$ and $p_2$, ${\rm Pr}\{p_1(\bm\beta_1 + \bm\alpha_2\bm\beta_2) + (\mbf{D} \bm\alpha_1 + \bm{e}_2)\bm\beta_2\} = {\rm Pr}\{p_2(\bm\beta_1 + \bm\alpha_2\bm\beta_2) + (\mbf{D} \bm\alpha_1 + \bm{e}_2)\bm\beta_2\}$, which simply yields $\bm\beta_1 = -\bm\alpha_2\bm\beta_2.$ Hence $\mbf{Y} = (\mbf{X}\bm - \mbf{P}\bm\alpha_2)\beta_2$.
\item Given the dataset, we estimate the parameters $\bm\alpha_2$ and $\bm\beta_2$ by partial least squares method, denoted the estimators as $\widehat{\bm\alpha}_2$ and $\widehat{\bm\beta}_2$. Then, the equation (2) can be obtained.  
\end{itemize}
Similar to equation (2), we can get equation (4). 
\begin{itemize}
  \item Intervene on $\mbf{Z}$ by removing all incoming arrows, see Figure 2, and set $\mbf{Z}=z'$, where $p'$ is a random variable. Then we obtain: 
\begin{align*}
{\mbf Z} = z',
{\mbf X} =  {\mbf D}{\bm \gamma}_1 +  {\mbf Z}{\bm\gamma}_2 + \bm{\epsilon}_2,
{\mbf Y} =  {\mbf Z}{\bm \theta}_1 +  {\mbf X}{\bm \theta}_2.    \end{align*}   
\item Integrate the first and second equations into the third equation from their structural equations. 
\begin{align*}
\mbf{Y} = z'(\bm\theta_1 + \bm\gamma_2\bm\theta_2) + \mbf{D} \bm\gamma_1\bm\theta_2 + \bm{\epsilon}_2\bm\theta_2.
\end{align*}
\item Require the distribution of $\mbf{Y}$ to be invariant under interventions $\mbf{D}$, i.e. for all $d_1$ and $d_2$, ${\rm Pr}\{z'(\bm\theta_1 + \bm\gamma_2\bm\theta_2) + d_1 \bm\gamma_1\bm\theta_2 + \bm{\epsilon}_2\bm\theta_2\} = {\rm Pr}\{z'(\bm\theta_1 + \bm\gamma_2\bm\theta_2) + d_2 \bm\gamma_1\bm\theta_2 + \bm{\epsilon}_2\bm\theta_2\}$, which simply yields $\bm\theta_2 = 0.$ Hence $\mbf{Y} = \mbf{Z}\bm\theta_1$.
\item Given the dataset, we estimate the parameter $\bm\theta_1$ by partial least squares method, denoted the estimator as $\widehat{\bm\theta}_1$. Then, equation (4) can be obtained.  
\end{itemize}




\end{document}